\documentclass[letterpaper]{article}
\usepackage{iccc}
\usepackage{times}
\usepackage{helvet}
\usepackage{courier}

\usepackage{comment}
\usepackage{graphicx} 
\usepackage{listings}
\usepackage{xcolor}
\usepackage{amsfonts}
\usepackage{amsmath}
\usepackage{graphicx}
\usepackage[bb=dsserif]{mathalpha}
\usepackage{bm}
\usepackage{float}
\usepackage{xurl} 
\usepackage{makecell} 
\usepackage{svg} 
\usepackage{xcolor}
\usepackage{booktabs}    
\usepackage{mathtools}
\usepackage{multirow}
\usepackage{caption}
\usepackage{subcaption}
\usepackage{tabularx}
\usepackage{amsmath,bm}
\usepackage{algorithm}
\usepackage{algpseudocode}
\usepackage{wrapfig}
\usepackage{lipsum}
\usepackage{tikz}
 
\usepackage{bm}

\usepackage{CJKutf8}        

\newcommand{\SpaceAfterFigure}{\vspace{-0.35em}}
\newcommand{\SpaceAfterCaption}{\vspace{-0.5em}}

\title{DiffCJK: Conditional Diffusion Model for High-Quality and Wide-coverage CJK Character Generation}
\author{Yingtao Tian\\Google DeepMind\\Tokyo, Japan\\\texttt{alantian@google.com}}
\date{January 2024}

\begin{document} 
\begin{CJK}{UTF8}{min}
\maketitle
\begin{abstract}

Chinese, Japanese, and Korean (CJK), with a vast number of native speakers, have profound influence on society and culture.
The typesetting of CJK languages carries a wide range of requirements due to the complexity of their scripts and unique literary traditions.
A critical aspect of this typesetting process is that CJK fonts need to provide a set of consistent-looking  glyphs for approximately one hundred thousand characters. 
However, creating such a font is inherently labor-intensive and expensive, which significantly hampers the development of new CJK fonts for typesetting, historical, aesthetic, or artistic purposes.

To bridge this gap, we are motivated by recent advancements in diffusion-based generative models and propose a novel diffusion method for generating glyphs in a targeted style from a \emph{single} conditioned, standard glyph form.
Our experiments show that our method is capable of generating fonts of both printed and hand-written styles, the latter of which presents a greater challenge.
Moreover, our approach shows remarkable zero-shot generalization capabilities for non-CJK but Chinese-inspired scripts.
We also show our method facilitates smooth style interpolation and generates bitmap images suitable for vectorization,
which is crucial in the font creation process. 
In summary, our proposed method opens the door to high-quality, generative model-assisted font creation for CJK characters, for both typesetting and artistic endeavors.

\end{abstract}

\begin{figure}[h!]
    \centering
    \includegraphics[width=0.85\columnwidth]{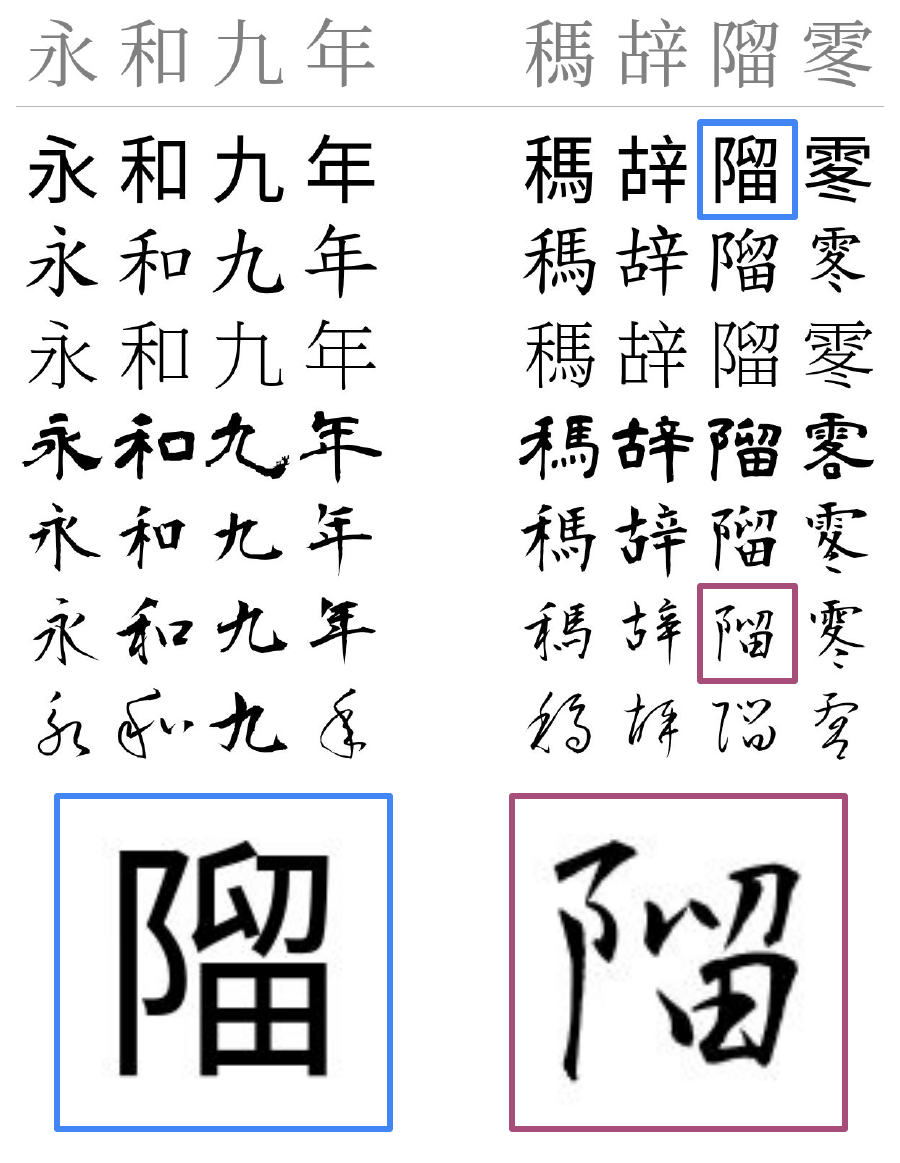}
    \SpaceAfterFigure{}
    \caption{Our method generates \emph{highly stylized} and \emph{legitimate} CJK glyphs,  For each character, our method refers to a standard font's bitmap (visualized in gray) and generates a diverse array of glyphs in various printed and calligraphy form (More details in Figure~\ref{fig:matrix}.) 
    Our method  is effective for both common (left) and extremely rare (right) CJK characters.
    The zoom-ins showcase examples of printed and calligraphy form, highlighting the method's high quality and its utility for font designers and artists alike.}
    \SpaceAfterCaption{}
    \label{fig:teaser}
\end{figure}

\section{Introduction}

The importance of East Asian languages that use Chinese characters, including Chinese, Japanese and Korean (CJK) languages, is profound. For example, they are used by more than $1.56$ billion speakers~\cite{authors2024ethnologue} which account for more than $25\%$ of global GDP~\cite{imf2023world}. Also, historically, a wide range of scripts have been heavily influenced by Chinese characters, adding more historical and culture value.
In the modern era, with the proliferation of printing and information technology, writing scripts need to address the challenges typesetting: for example, the requirements on the visual appearance of characters means that a font must provide well-designed and consistent-looking glyphs across the glyph repertoire.

However, the unique aspect for CJK typesetting is the sheer number of characters that makes such requirement demande more effort: CJK contains nearly one-hundred thousand unique characters in the latest Unicode 15.1~\cite{unicode2023v15dot1}. 
Furthermore, there is also an emphasis on typefaces categories with different styles\footnote{Here ``style'' is in a general sense rather than its specific meaning in font development as in ``style instance''.}, as they carry practical, cultural, and artistical meaning~\cite{huang2020type,kim2020typejp,kim2020typekr}. 
Consequently, the font industry faces a dilemma: 
either making one style instance with as wide of coverage as possible, 
or making 
fonts with wider style support
but with limited coverage.
Despite these compromises, font creation remains an laborious and expensive task.
Therefore, it remains a critical challenge to efficiently producing fonts that encompass a diverse range of \emph{various, multi-style} yet \emph{legitimate} CJK glyphs for the entire set of nearly one hundred thousand characters.

To address these challenges, we propose a diffusion-based method (Figure~\ref{fig:teaser}) capable of generating characters in diverse styles conditioned on a \emph{single} standard reference glyph. 
The reference glyph only needs to outline the character's shape, obviating the need for intricate design work. 
As a diffusion model, it consists of two processes: a diffusion process that gradually adds noise to destroy the data, and a reverse process that generates new samples from scratch by progressively removing noise. 
A deep neural network is trained to estimate the noise, conditioned on the reference character.
Unlike common text-to-image models which must memorize the shapes of characters and thus suffering from limited data of rare characters, our method works uniformly well for all CJK glyphs, from common to rare ones, as long as we have an available reference.
Furthermore, by injecting style information in the temporal embedding for diffusion,
our method can not only generate characters in various styles but also interpolate between these styles.

Experiments show that our method can generating legitimate CJK characters across a broad spectrum of styles.
This includes typefaces for physical and digital typesetting, as well as artistic calligraphy styles. 
The later poses a greater challenge due to more limited data and significant stylistic deviations from reference glyphs. 
Moreover, our method achieves zero-shot generalization to CJK-inspired scripts not encountered in training (such as the under-resourced Cho Num and Tangut scripts)
and can meaningfully interpolate between styles. 
These capabilities enable font designers to produce fonts on a large scale, even with limited resources.
Additionally, our analysis highlights the efficiency of our method and its potential for vectorization, indicating its practicality for adoption and adaptation in font design workflows.

\section{Background and Related Works}

\subsection{Chinese Character in CJK Typesetting}

Chinese characters constitute a logographic writing system, circa 13th century BCE to present, utilized across multiple languages within the Sinosphere, or the East Asian cultural sphere~\cite{marginson2011higher}. Initially developed for Chinese, it has also been adopted for Japanese, Korean, Vietnamese, and has significantly influenced lesser-known, historical scripts like the Tangut and Khitan small scripts.

\begin{figure}[t!]
    \newlength{\subcharwidth}
    \setlength{\subcharwidth}{2.3em}
    \setlength{\tabcolsep}{5pt}
    \begin{tabular}{cccccc}
        \includegraphics[width=\subcharwidth]{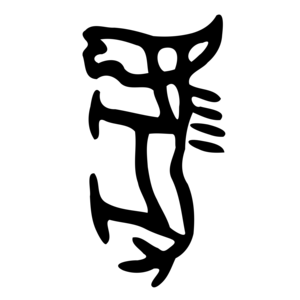} & \includegraphics[width=\subcharwidth]{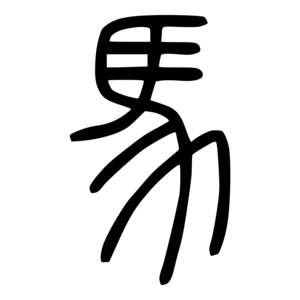} & \includegraphics[width=\subcharwidth]{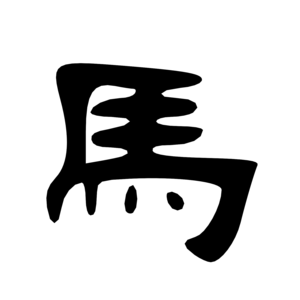} & \includegraphics[width=\subcharwidth]{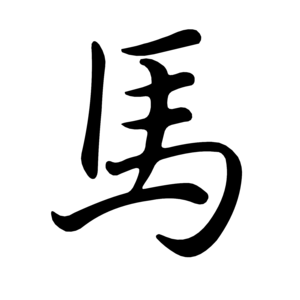} & \includegraphics[width=\subcharwidth]{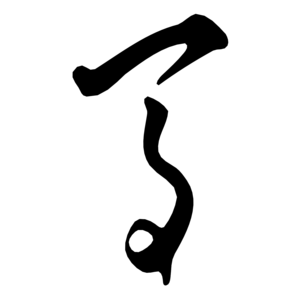} & \includegraphics[width=\subcharwidth]{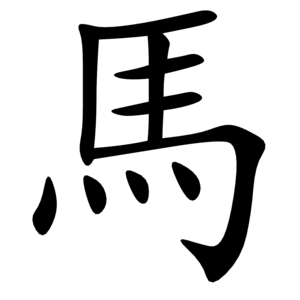} \\
        \hline
        \makecell{\small{} Oracle bone \\ script \\ (甲骨文)} & \makecell{\small{}Seal \\ script \\ (篆書)} & \makecell{\small{}Clerical \\ script \\ (隸書)} & \makecell{\small{}Semi-\\cursive \\ script \\ (行書)} & \makecell{\small{}Cursive \\ script \\ (草書)} & \makecell{\small{}Regular \\ script \\ (楷書)} \\
    \end{tabular}
    \SpaceAfterFigure{}
    \caption{Example of hand-written Chinese script styles for ``馬'' (horse)~\protect\cite{wiki:commonsancient}.}
    \SpaceAfterCaption{}
    \label{figure_example-of-script-styles}
\end{figure}

\begin{figure}[t!]
    \centering
    \begin{tabular}{c}
        \begin{subfigure}[b]{\columnwidth}
            \centering
            \includegraphics[width=0.85\columnwidth]{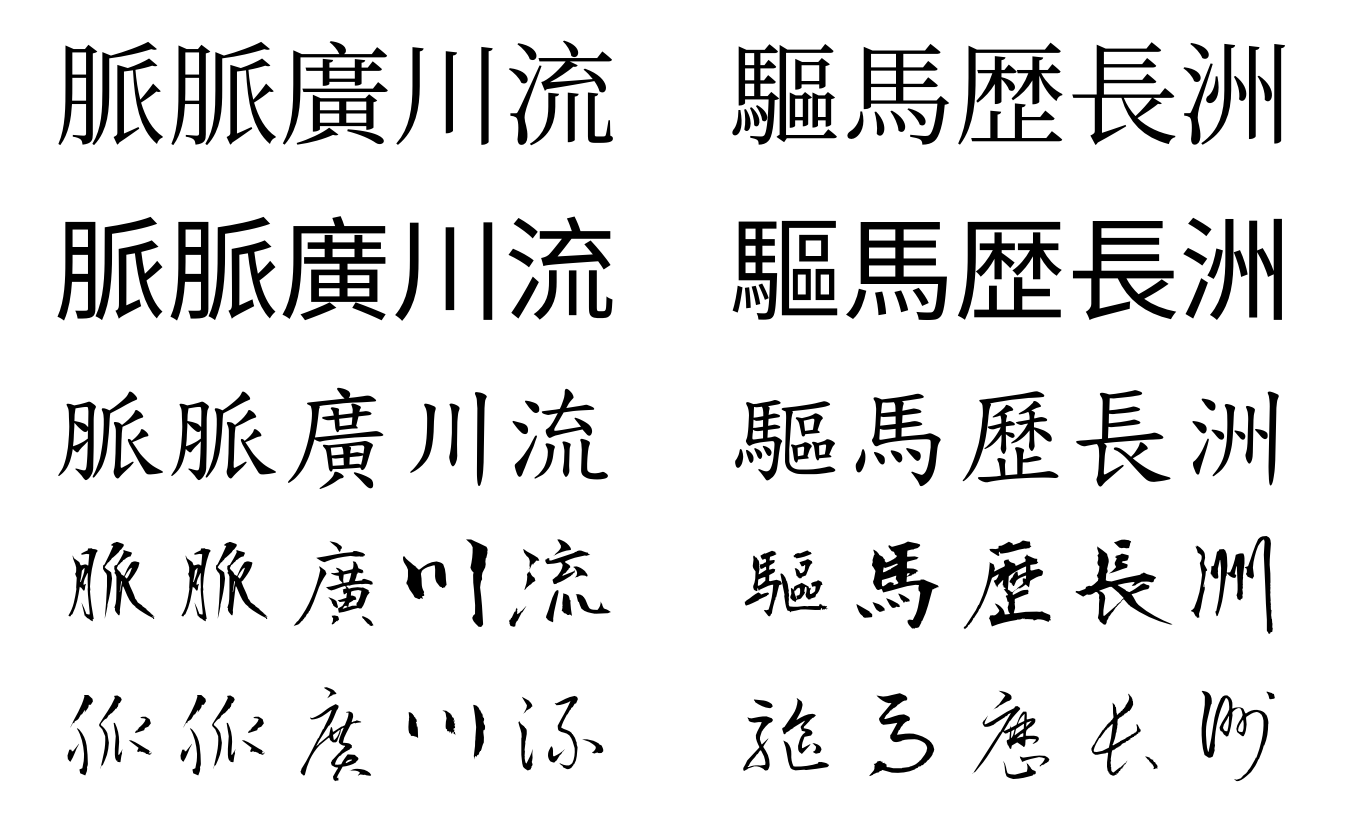}
        \end{subfigure} \\
        \begin{subfigure}[b]{\columnwidth}
            \centering
            \includegraphics[width=0.85\columnwidth]{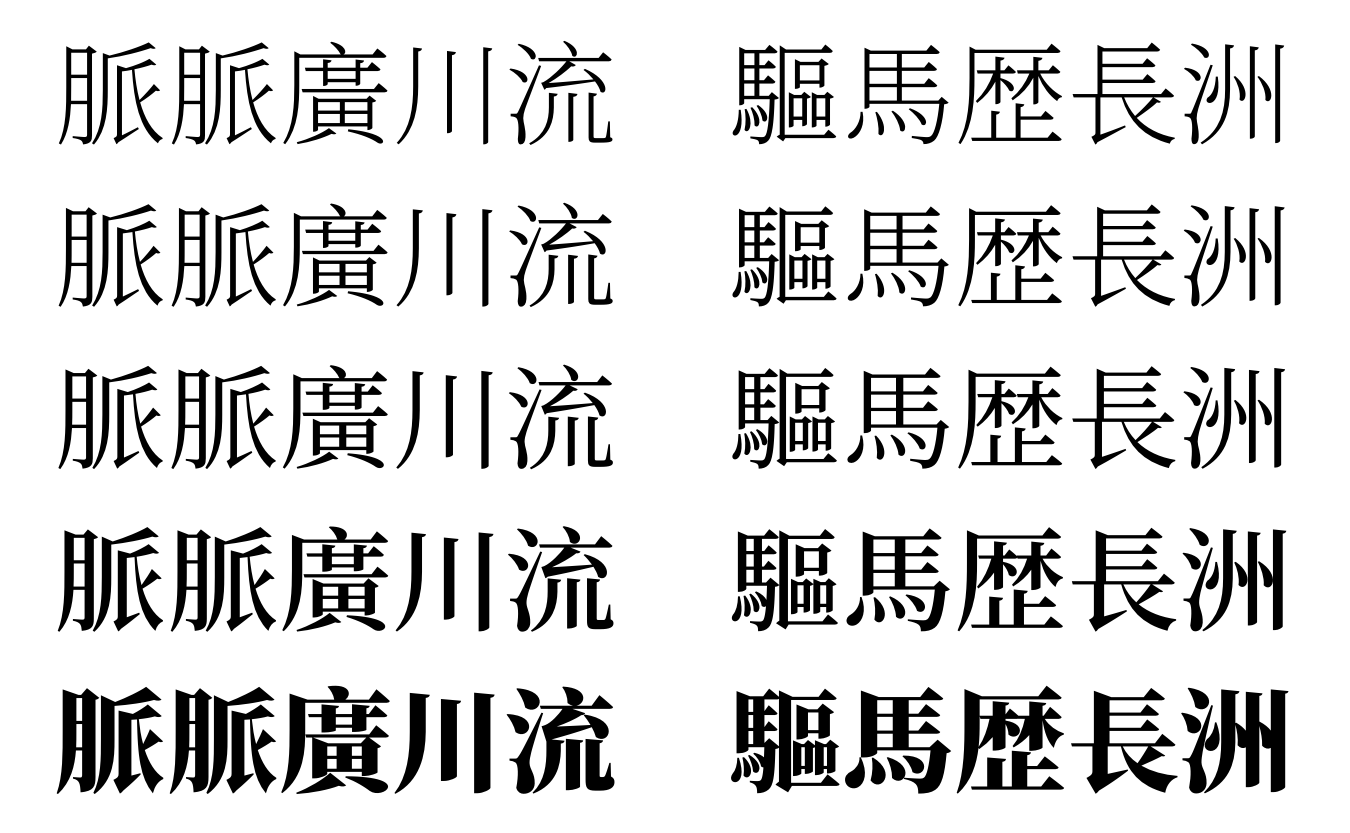}
        \end{subfigure}
    \end{tabular}
    \SpaceAfterFigure{}
    \caption{Example of CJK typefaces (above) and weights (below). \textbf{Typefaces}: (1) \emph{Ming} a.k.a. \emph{Song}, (2) \emph{Gothic typefaces}, (3) \emph{Regular script}, (4) \emph{Semi-cursive} and (5) \emph{cursive} script.
    \textbf{Widths}: Noto Serif CJK of different width: \emph{ExtraLight}, \emph{Light}, \emph{Normal}, \emph{SemiBold} and \emph{Black} (i.e.~\emph{Bold}.) }
    \SpaceAfterCaption{}
    \label{fig:example-of-cjk-typeface-and-weight}
\end{figure}

Like for any writing system, both physical~\cite{needham1974science} and digital~\cite{liu2022deciphering} typesetting has a profound impact in East Asian culture. 
In such typesetting practices, alongside \emph{how characters are arranged}, an equally important aspect is \emph{what characters look like}~\cite{w3jlreq,w3clreq,w3klreq} which necessitates \emph{glyphs} that are legitimate, consistent, and available for the vast number of characters.
This requirement is deeply rooted in pre-typesetting traditions, particularly the established script styles for handwriting. 
As shown in Figure~\ref{figure_example-of-script-styles}, 
there are five major script styles --- \emph{seal}, \emph{clerical}, \emph{semi-cursive}, \emph{cursive} and \emph{regular} style --- traditionally used to write Chinese characters, in additional to oracle bone script, the oldest known form of Chinese characters.
These styles naturally transitioned into typesetting~\cite{authors20xxtype} practice which will be elaborated below.
In the context of typesetting, the term \textbf{CJK} characters is used to collectively describe the glyphs for Chinese, Japanese, and Korean languages, all of which incorporate Chinese characters into their writing systems. 
Technically, CJK also encompasses derivatives of Chinese characters, such as the Kana scripts (Hiragana and Katakana) used in Japanese, although these are beyond the scope of this paper.

The importance of glyphs in typesetting is multifaceted, with style (typefaces) and weight~\cite{elliot2020introducing,elliot2020making} being two crucial aspects.
In Latin typesetting,  well-recognized styles include the serif (``brown fox'') and sans-serif ({\fontfamily{cmss}\selectfont ``brown fox''}), while weight variations are exemplified by bold (\textbf{``brown fox''}) and italics (\textit{``brown fox''}). 
CJK typesetting is not an exception, and even encompasses more delicacy regarding this matter:
it requires a consistent appearance across a vast number of characters combined with unique cultural practices.
Also, CJK typesetting places a significant emphasis on typeface categories, which carry practical, cultural, and artistic significance~\cite{huang2020type,kim2020typejp,kim2020typekr}.

Notably, styles in CJK typesetting, while similar, do not entirely align with Latin typesetting practices.
They are better shown by several key styles unique to CJK typography in Figure~\ref{fig:example-of-cjk-typeface-and-weight}: 
(1) \emph{Ming}/\emph{Song}, a printed form style that has evolved from the regular script over centuries, is the most widely used and is akin the serif style in Latin typography.
(2) \emph{Gothic typefaces}, printed form styles that convey a sense of modernity and are akin to sans serif styles in Latin typography.
(3) \emph{Regular} script, a printed form style that mimics handwritten forms, is primarily used for educational purposes and also serves a role similar to italics in Latin typography, although italics per se do not exist in CJK typesetting.
(4) Calligraphy form, such as \emph{semi-cursive} and \emph{cursive} script, that are rarely used in standard typesetting  but are favored for artistic expression.
Regarding font width, there exists a broad spectrum from light to black (which is akin to bold in Latin typography).
However, adjusting font weight in CJK characters involves more than merely altering stroke width: The exact boldness of each stroke must be carefully tuned to ensure visual consistency across characters with varying stroke counts, thus requiring more effort.

For the sake of completeness, we also want to emphasize the regional difference of CJK characters~\cite{unicodetechnicalnote26}.
These regional variations in typesetting are handled by both font design and encoding.
Another point of distinction is the use of simplified versus traditional characters,  which are represented by different code points. However, these aspects, while important, fall outside the scope of this paper and and we leave them for further study.

\begin{figure}[t!]
    \centering
    \includegraphics[width=0.8\columnwidth]{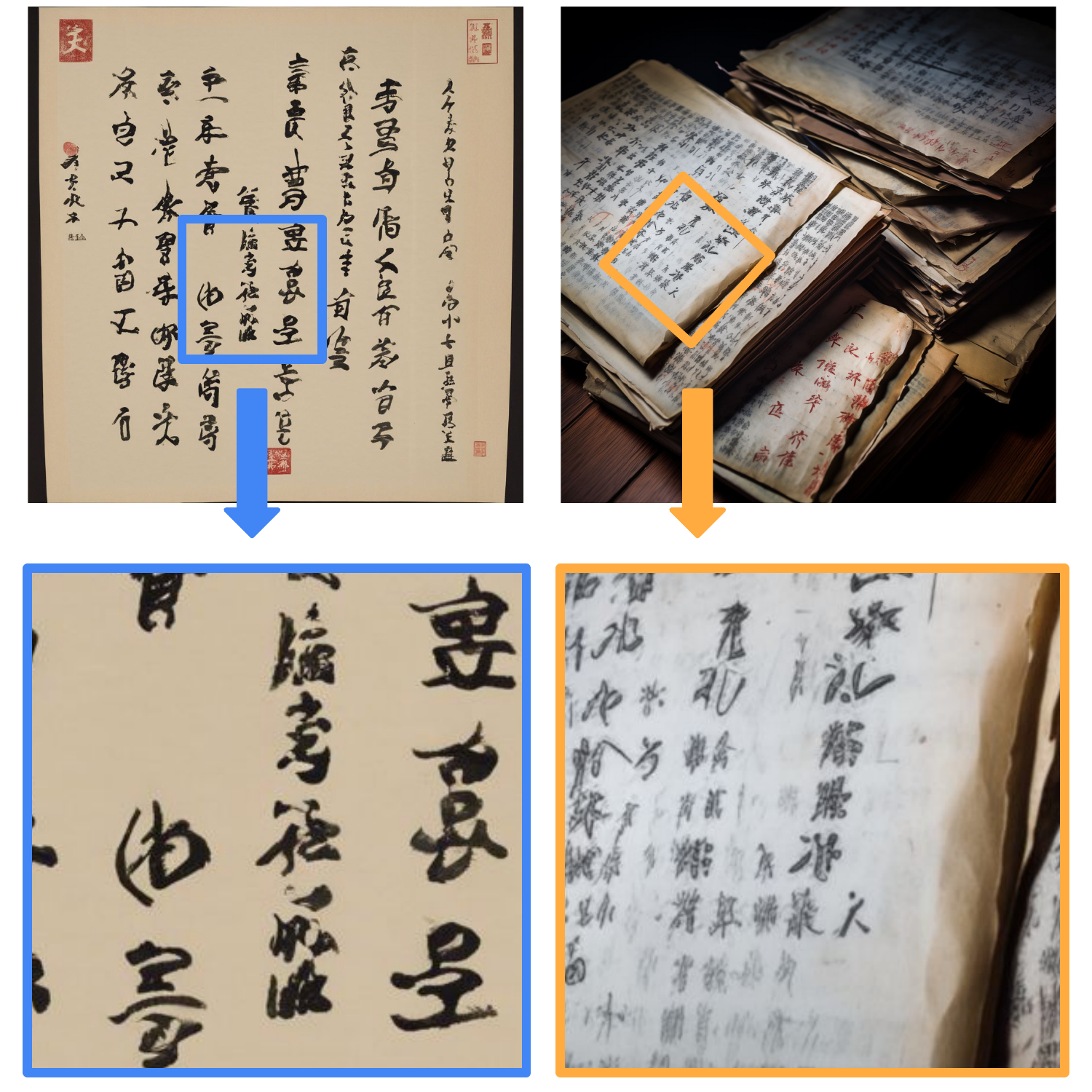}
    \SpaceAfterFigure{}
    \caption{Example of generating CJK characters using state-of-the-art text-to-image models. From left to right: Stable Diffusion XL~\protect\cite{podell2023sdxl}, MidJourney v6~\protect\cite{authors2024midjourney},
    and zoomed-in views are on the lower row.
    While these models are powerful and expressive, the generated characters are illegitimate to native speakers.}
    \SpaceAfterCaption{}
    \label{fig:example-of-text-to-image}
\end{figure}

\begin{figure}[h!]
    \centering
    \includegraphics[width=0.75\columnwidth]{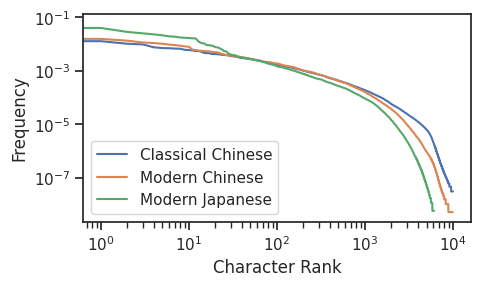}
    \SpaceAfterFigure{}
    \caption{Distribution of CJK characters in Classical Chinese, Modern Chinese and Modern Japanese. 
    Besides frequent characters, most of the characters are under-represented in the text data.
    This plot uses data aggregated from corpus statistics~\protect\cite{jun2010chinese,ninjal2015}.}
    \SpaceAfterCaption{}
    \label{fig:distribution-of-cjk-characters}
\end{figure}

\subsection{Computation and Creative Approach to CJK Glyphs}

The advances of generative models have led to image generation models capable of producing high quality images comparable to professional photography and artwork. 
Noteworthy examples of such models that are public-available and state-of-the-art include Stable Diffusion XL~\protect\cite{podell2023sdxl} and MidJourney v6~\protect\cite{authors2024midjourney}.
The former is open-source while the later is offered as a commercial product. 
Given their powerful capabilities, it is natural to apply these tools for producing CJK glyphs.
However, generating valid CJK characters turns out to remain a challenging task even for these advanced models. 
As illustrated in Figure~\ref{fig:example-of-text-to-image}, 
these state-of-the-art image generation models fail to create legitimate CJK characters despite their ability to produce valid scenes and fine-grain details.

\begin{figure*}[th]
    \centering
    \includegraphics[width=0.95\textwidth]{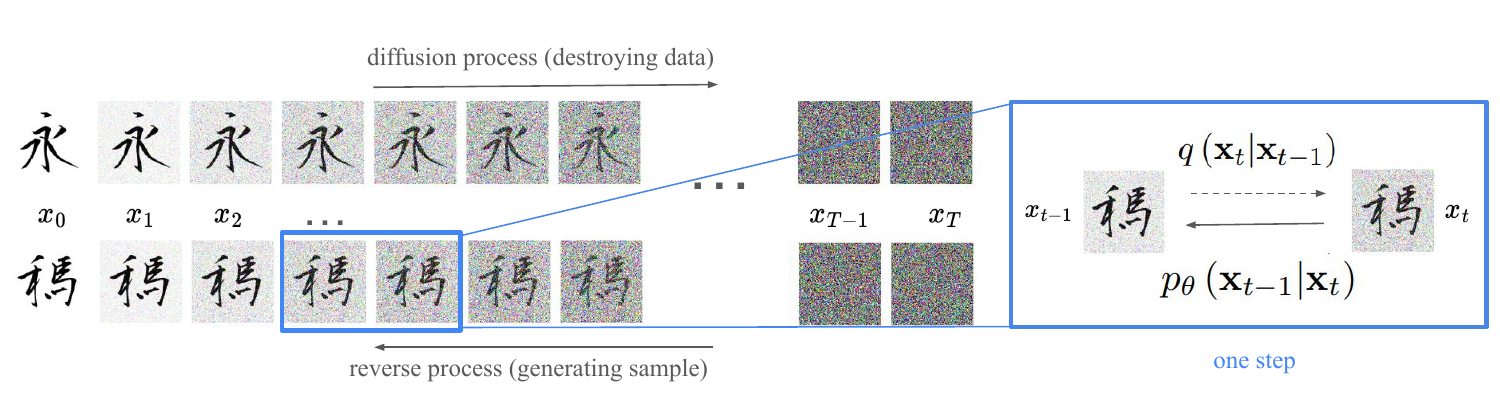}
    \SpaceAfterFigure{}
    \caption{The diffusion model consists of two processes:  (1) the diffusion process that gradually adds noise to destroy the data, and (2) the reverse process that gradually removes the noise through a Markov process (sampling from $p_{\theta} \left( \bm{x}_{t-1} | \bm{x}_t \right)$ at step $t-1$) for generating new samples. The estimation of the noise is parameterized by a deep learning model. }
    \SpaceAfterCaption{}
    \label{fig:arch-diffusion}
\end{figure*}

For text-to-image models, semi-supervised training is important for generating high-fidelity images~\cite{zhou2023shifted}.
However, we argue that the fundamental issue with the inability of state-of-the-art model to produce legitimate CJK characters lies primarily in the very distribution of characters.
Specifically, the challenges are two fold, involving both the amount and complexity of CJK characters:
Firstly, regarding number of characters, the latest version of Unicode, Unicode 15.1, encodes a total of \textbf{97,680} characters.
Secondly, a Figure~\ref{fig:distribution-of-cjk-characters} show, the frequency distribution of characters, which follows power-law, suggests that only a small subset of characters appears frequently enough to enable effective learning by any model.
As a result, a vast number of CJK characters remains underrepresented, introducing difficulty for any model dealing with them.

Another line of work in generating CJK characters involves transferring from a standardized glyph. 
Specifically, it has been proposed to generate the glyph in the desired style based on a standard glyph, usually using a reference font that covers CJK code points comprehensively.
This approach becomes promising since only a small number of fonts covers a complete set of CJK code points due to significant effort required to produce a consistent appearance for all CJK characters.
Examples of such reference fonts that are free to use include Noto CJK~\cite{notofonts} and Jigmo~\cite{jigmofonts}.
The availability of these comprehensive fonts enables the training of a conditional generative model on a limited set of paired glyphs, which can be applied to the entire set of CJK in Unicode. 
Many of these models are based on Generative Adversarial Networks (GAN)~\cite{zhang2018separating,tian2018zi2zi,gao2019artistic,lo2019font2font,zhu2020few,park2021few,liu2023fonttransformer}.
Some of them leverage additional information about components and composition.
While these efforts has seen promising results, they still face quality issues observable upon closer inspection, likely due to the inherent challenges of training GANs.

Perhaps the work that is mostly closely related to ours, in that it uses a Denoising Diffusion Probabilistic Model (DDPM)~\cite{ho2020denoising} is \cite{gui2023zero}, which introduces Glyph Conditional DDPM (GC-DDPM) that is built on a UNet architecture~\cite{ronneberger2015u} and generates glyphs from a reference glyph.
The key distinction lies in that our method focuses on both typesetting style and brush-based calligraphy generation for font designing and artist, aiming at publisher-level quality suitable for professional font design.
In contrast, GC-DDPM aims to synthesize data to enhance the performance of  downstream classification on stroke-based hand-writing recognition tasks through data augmentation.
Nonetheless,  the design decisions made in GC-DDPM have provided valuable insights for our approach.

\subsection{Diffusion Model}

Diffusion models~\cite{sohl2015deep,ho2020denoising,nichol2021improved,song2020score} represent a broad spectrum of deep generative models that have established the state-of-the-art in a wide range of challenging tasks.
A very limited set of examples from the large body of diffusion works includes computer vision~\cite{nichol2021improved,song2020score}, image synthesis~\cite{ruiz2023dreambooth,authors2024midjourney,podell2023sdxl}, natural language processing~\cite{lovelace2024latent,wu2024ar}, and signal processing~\cite{engel2020ddsp,goel2022s}.
Given the expansive growth and wide applications, this  emerging field is best navigated through comprehensive surveys~\cite{yang2023diffusion,croitoru2023diffusion}.
The specifics of how we leverage diffusion models are elaborated upon in the subsequent section.

\begin{figure}
    \centering
    \includegraphics[width=0.95\columnwidth]{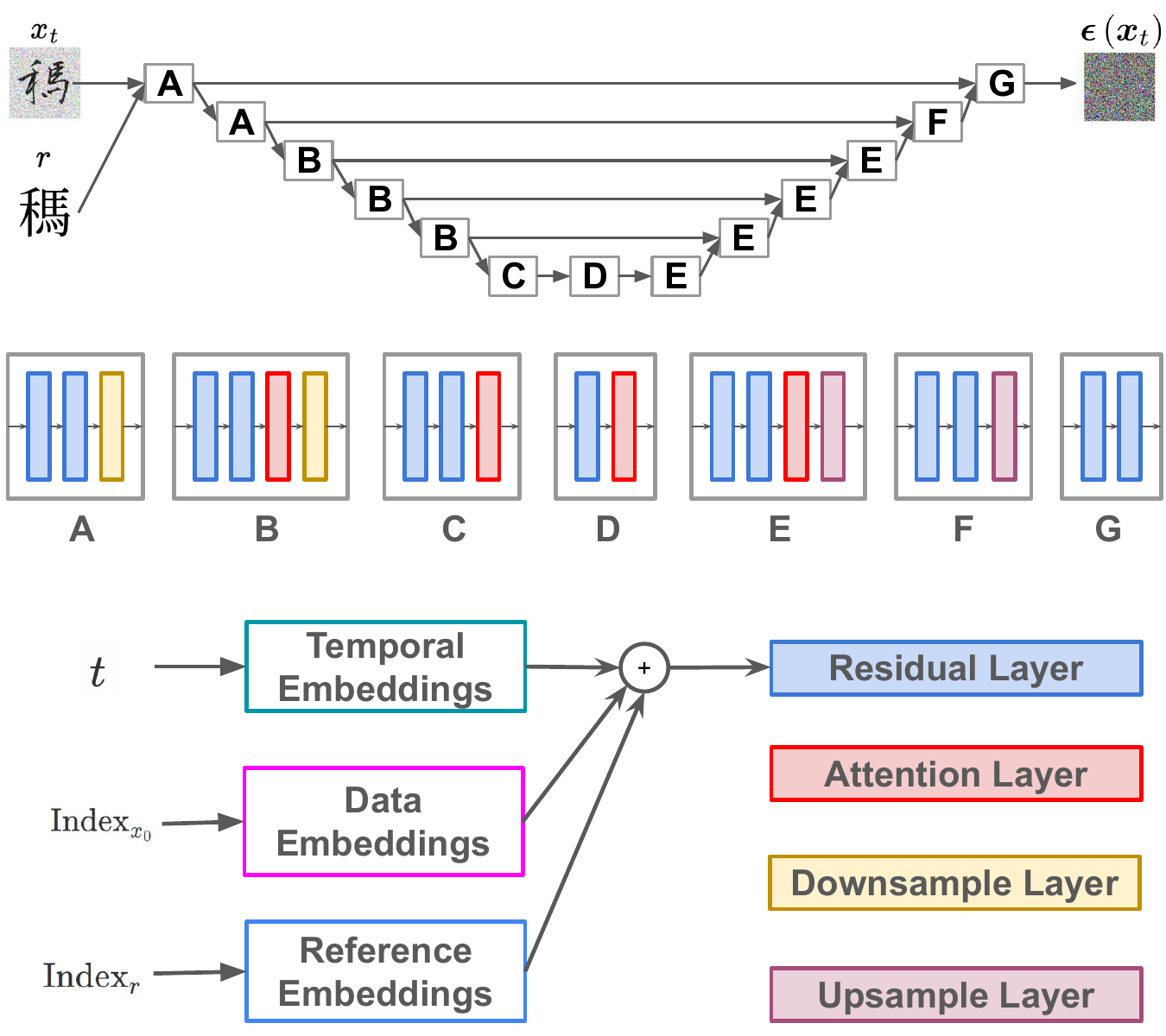}
    \SpaceAfterFigure{}
    \caption{The UNet used in our proposed method.}
    \SpaceAfterCaption{}
    \label{fig:arch-net}
\end{figure}

\section{Method}

Our proposed method employs a diffusion model~\cite{sohl2015deep} and we adopt the notation used DDPM~\cite{ho2020denoising} here.
As illustrated in Figure~\ref{fig:arch-diffusion}, the diffusion model consists of two processes.
Denoting data as $\mathbf{x}_0 \sim q(\mathbf{x_0})$,
the diffusion process, also known as the forward process, gradually adds noise to destroy data $\mathbf{x}_0$, producing a sequence of progressively more noisy samples $\mathbf{x}_1, \mathbf{x}_2, \cdots$ until the entirely noisy sample $\mathbf{x}_T$.
Given a schedule of variance, the forward process is a Markov process $\mathbf{x}_t \sim q\left( \mathbf{x}_t | \mathbf{x}_{t-1} \right) = \mathcal{N} \left( \mathbf{x}_t; \sqrt{1 - \beta_t} \mathbf{x}_{t-1}, \beta_t \mathbf{I} \right)$, and approximates the posterior.

On the other hand, the reverse process gradually removes the noise (``denoising'') through a Markov process $\mathbf{x}_{t-1} \sim p_{\theta} \left( \mathbf{x}_{t-1} | \mathbf{x}_t \right) = \mathcal{N}\left( \mathbf{x} ; \bm{\mu}_{\theta}\left( \mathbf{x}, t\right), \mathbf{\Sigma}_{\theta}(\left( \mathbf{x} ,t\right) \right) $. 
As the reverse process approximates the prior of data $\mathbf{x}_0$, it can be used for generating samples.
Note that if $\beta_t$ is small enough, $q\left( \mathbf{x}_t | \mathbf{x}_{t-1} \right)$ would be Gaussian, so we have $q\left(\mathbf{x}_t | \mathbf{x}_0 \right) = \mathcal{N}\left( \mathbf{x}_t; \sqrt{\overline{\alpha}_t} \mathbf{x}_0, (1-\overline{\alpha}_t), \mathbf{I}\right)$ where $\overline{\alpha} = \prod_{s=1}^t{\alpha_s}$ and $\alpha_t = 1 - \beta_t$.
This means we can approximate  $q$ in the following way:
$q\left(\mathbf{x}_{t-1} | \mathbf{x}_t, \mathbf{x}_0 \right) = \mathcal{N}\left(  \mathbf{x}_{t-1};  \tilde{\bm{\mu}}_{t} \left(\bm{x}_t,\bm{x}_0\right),\tilde{\beta}_t \bm{I} \right)$ where
$   \tilde{\bm{\mu}}_{t} \left(\bm{x}_t,\bm{x}_0\right)     =    \frac{\sqrt{\overline{\alpha}_{t-1}}\beta_t}{1-\overline{\alpha}_t} \bm{x}_0 + \frac{\sqrt{\alpha_t}\left(1-\overline{\alpha}_{t-1}\right)}{1-\overline{\alpha}_t}\bm{x}_t$ and $\tilde{\beta}_t = \frac{1-\overline{\alpha}_{t-1}}{1-\overline{\alpha}_t}\beta_t$.

The training in practice has a few notable realizations as suggested in the DDPM literature~\cite{ho2020denoising,nichol2021improved}.
First, the training process can sample arbitrary $t$ instead of going from $T$ to $1$.
Also, a deep neural network is tasked with predicting the noise added to $\bm{x}_0$ from partially corrupted $\bm{x}_t$, since we have
$\bm{x}_{0} = \frac{1}{\sqrt{\alpha_t}} \left(  \bm{x}_t - \frac{\beta_t}{\sqrt{1-\overline{\alpha}_t} } \bm{\epsilon}\left(\bm{x}_t\right)   \right)$.
We follow the suggestions~\cite{ho2020denoising} that the network should predict $\bm{\epsilon}\left(\bm{x}_t\right)$.
Although, it is recommended by~\cite{nichol2021improved} DDPM can be improved with learned variance, our empirical findings suggest that the base DDPM model's performance is sufficiently robust, and gains from such modifications could be minimal.

For the network architecture, we train a U-Net~\cite{ronneberger2015u} to predict $\bm{\epsilon}\left(\bm{x}_t\right)$ as shown in Figure~\ref{fig:arch-net}.
The overall network structure is a typical UNet similar to that of GC-DDPM~\cite{gui2023zero}.
This UNet, taking $\bm{x}_t$ as the input and predicting the noise  $\bm{\epsilon}\left(\bm{x}_t\right)$, consists of several blocks that first downsample and then upsample.
These blocks are connected by skip-connections. 
Each block is composed of residual, attention, downsample and/or upsample layers.
To allow controlling of the generation, we feed the reference image $r$ by concatenating it with $\bm{x}_t$. Also, the timestep $t$, index of $\bm{x}_0$, and $r$ are injected through embeddings.
After training, we can sample images by iteratively reducing the noise predicted by the U-Net.

\section{Experiments}

\subsection{Experiment Setup}

\begin{table}[t]
    \begin{tabular}{lrl}
        \hline
        Style                     &  \# / Form $^a$  & Font               \\ \hline
        {Serif (Ming/Song)}       & 43062 / P & Noto Serif TC      \\ \hline
        {Gothic Typefaces}        & 43098 / P & Noto Sans TC       \\ \hline
        {Regular Script} $^b$     & 83534 / P & TW-Kai             \\ \hline
        {Serif (Ming/Song)} $^b$  & 83534 / P & TW-Sung            \\ \hline
        {Clerical Script}         & 7349 / C  & AoyagiReisho       \\ \hline
        {Semi-cursive Script }    & 8865 / C  & KouzanMouhitu      \\ \hline
        {Semi-cursive Script}     & 7360 / C  & KouzanGyousho      \\ \hline
        {Cursive Script}          & 7741 / C  & KouzanSousho       \\ \hline
        {Serif (Ming/Song)} $^c$  & 50217 / P & Noto Serif Tangut  \\ \hline
        {Serif (Ming/Song)} $^d$  & 22741 / P & NomNaTong          \\ \hline
        \multicolumn{3}{l}{\begin{tabular}[c]{@{}l@{}}
        $^a$ P for Printed From and C for Calligraphy Form.\\$^b$ CNS 11643 Standard.
        $^c$ Only for Tangut Script.\\ $^d$ Contains CJK, including Chu Nom
        \end{tabular}} \\ 
    \end{tabular}
    \SpaceAfterFigure{}
    \SpaceAfterCaption{}
    \caption{Fonts we used for training and/or inference. We focuses on two common types in CJK: typefaces for typesetting and calligraphy for artistical purpose.  All fonts are open fonts that are free to use.}
    \label{tbl:fonts}
\end{table}

\subsubsection{Dataset}

For training and inference, we use publicly-available fonts since they provide a large set of glyphs with a consistent look. 
Concretely, we use a wide range of free fonts in both printed and calligraphy forms, detailed in Table~\ref{tbl:fonts}.
During training, we construct glyph pairs by looking at the the shared glyphs between the two fonts.

\subsubsection{Model and Training Details}

Our network configuration is detailed as follows: each blocks in Figure~\ref{fig:arch-net} has respectively $128$, $128$, $256$, $256$, $512$, $512$, $512$, $512$, $512$, $256$, $256$, $128$, $128$ channels respectively in their residual layers. 
All embeddings utilized are uniformly set to $512$ dimensions.
The network accepts inputs with spatial dimensions of $128$ by $128$, which are halved in each downsample block, reaching a minimum dimension of $2 \times 2$ pixels at block $D$.

For training, Noto Serif TC is employed as the reference font, with all other fonts serving as target fonts for the model to learn from.
The training process contains $3000$ epochs, and spans a total duration of one week with 16 A100 GPUs.

\begin{figure*}[t]
    \centering
    \includegraphics[width=0.925\textwidth]{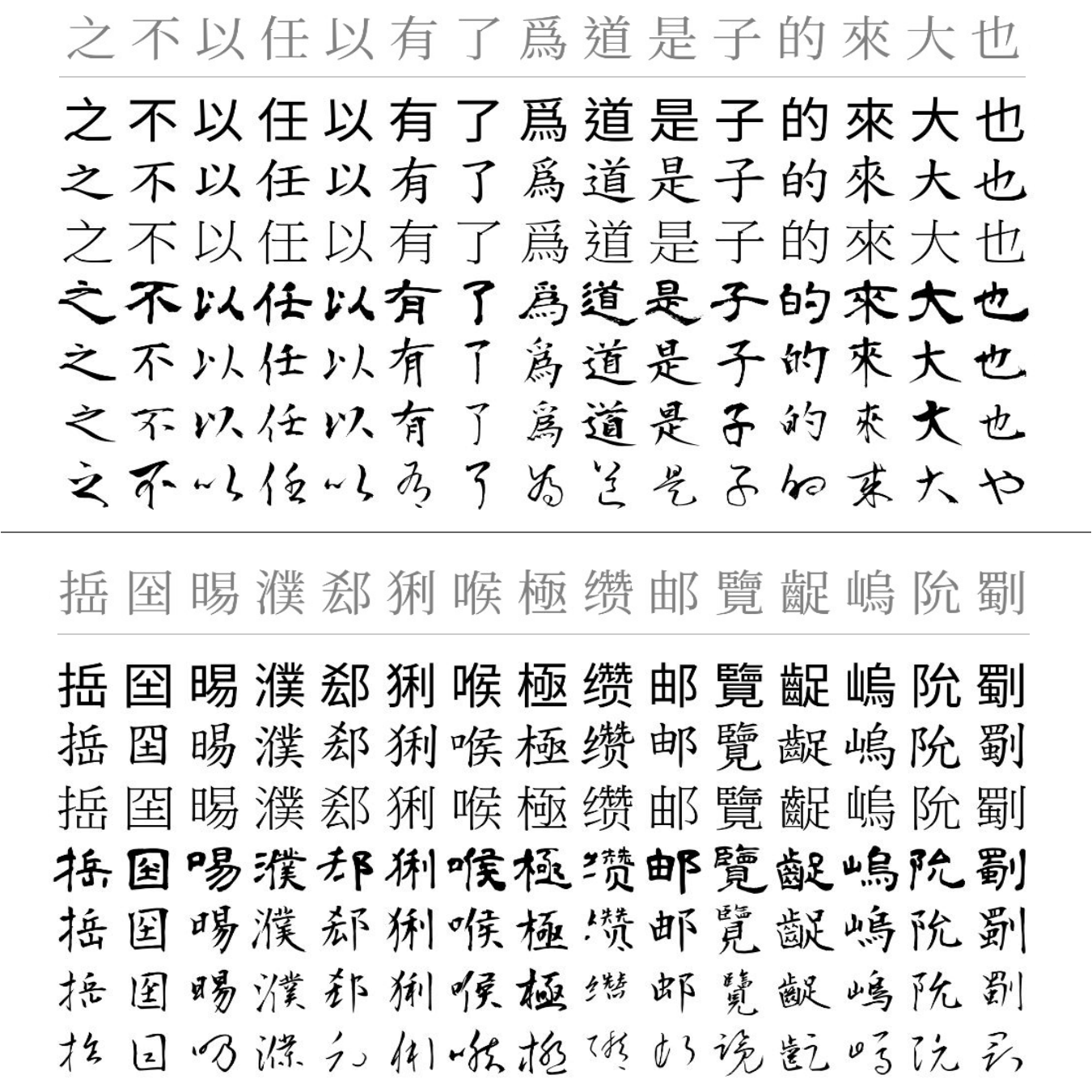}
    \SpaceAfterFigure{}
    \caption{Our method generating CJK characters in a wide range of styles. The upper block is very-commonly-used characters, while the lower block are characters drawn randomly from a collection of less common ones. In each block, we show reference characters  in gray, and generated characters in black. Each row shows different styles our method could generate: they are respectively (1) \emph{Gothic typefaces}, (2) \emph{Regular script}, (3) \emph{Song/Ming} in a different standard (CNS 11643), (4) \emph{Clerical script}, (5-6) two different \emph{Semi-cursive scripts} and (7) \emph{Cursive script}. See Table~\protect\ref{tbl:fonts} for font data.}
    \SpaceAfterCaption{}
    \label{fig:matrix}
\end{figure*}

\subsection{Overview of Generation Capability}

To demonstrate the overall generation capability of our proposed method, we present a matrix of generated results in Figure~\ref{fig:matrix}, featuring a wide range of fonts from Table~\ref{tbl:fonts}.
Our method applies to both common and rare characters, the latter not present in the training dataset.
Specifically, we using Noto Serif TC, a Serif (\emph{Ming/Song}) font as the reference to generate characters in the desired fonts.
These desired fonts include a wide range of styles such as (1) Noto Sans TC (\emph{Gothic Typefaces}), (2) TW-Kai (\emph{Regular Script}), (3) TW-Sung (\emph{Serif (Ming/Song)}), along with calligraphy styles including (4) AoyagiReisho (\emph{Clerical Script}), (5-6) KouzanMouhitu and KouzanGyousho (\emph{Semi-cursive Script}), (7) KouzanSousho (\emph{Cursive Script}). 
The selection of common characters is based on statistics~\cite{jun2010chinese}, while less common characters are randomly sampled from Unicode block \emph{CJK Unified Ideographs}~\cite{unicode2023v15dot1}, which contains the first batch of $20,992$ CJK characters in Unicode.
It is shown that our proposed method consistently generates high-quality characters across this diverse range of styles, which are coherent and recognizable to native speakers.

\subsection{Converting between Printed and Calligraphy Form}

\begin{figure}[t!]
    \centering
    \includegraphics[width=0.85\columnwidth]{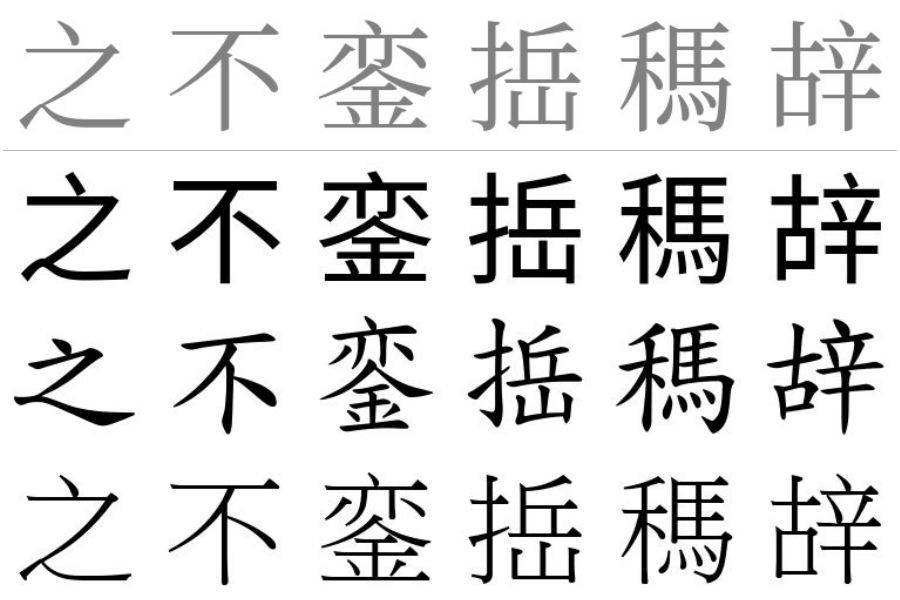}
    \SpaceAfterFigure{}
    \caption{Generating Printed form. The generated characters are in black, and three rows show \emph{Gothic typefaces}, \emph{Regular script}, \emph{Song/Ming} in a different standard (CNS 11643) respectively. First two characters are most common, middle two less common and final two extremely rare as detailed in the main text.}
    \SpaceAfterCaption{}
    \label{fig:typefaces}
\end{figure}

Here we explore the model's ability to generate printed form glyphs, focusing on the subtle distinctions found in fine-grain details. As shown in Figure~\ref{fig:typefaces}. we demonstrate generating \emph{Gothic typefaces}, \emph{Regular script} and \emph{Song/Ming} with different characters: first two are most common characters from statistics on Classical Chinese, middle two are less common ones from Unicode block \emph{CJK Unified Ideographs}, and last two are extremely rare ones sampled from Unicode block \emph{CJK Unified Ideographs Extension B}~\cite{unicode2023v15dot1} that contains $42,720$ characters extremely rare and historical characters.
Our method is capable of accurately generating characters in a variety of styles, and works for a wide range of characters uniformly.

\begin{figure}[h!]
    \centering
    \includegraphics[width=0.85\columnwidth]{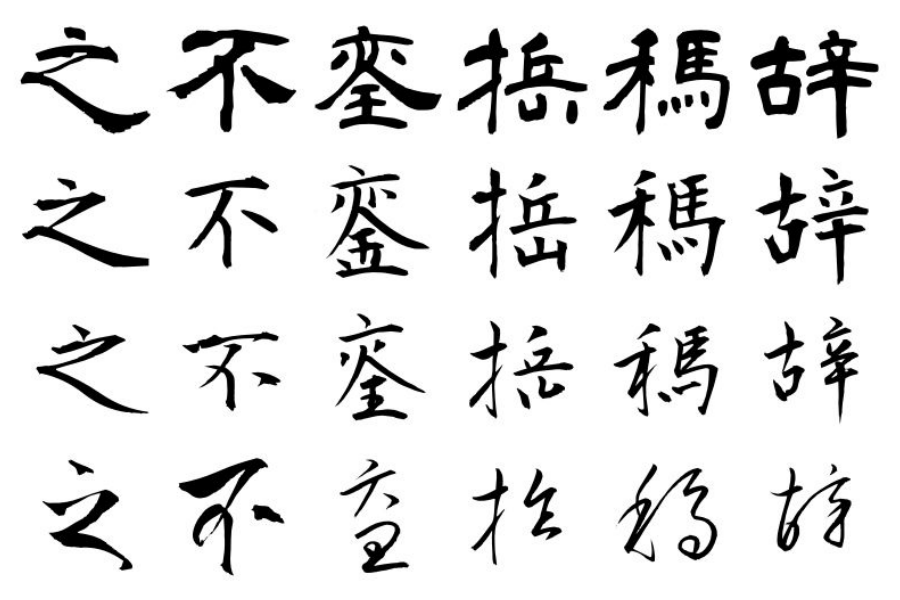}
    \SpaceAfterFigure{}
    \caption{Generating the same set of characters in Figure~\ref{fig:typefaces} in (4) \emph{Clerical script}, (5-6) two examples of \emph{Semi-cursive script} and (7) \emph{Cursive script}.}
    \SpaceAfterCaption{}
    \label{fig:caligraphies}
\end{figure}

Furthermore, we examine our proposed method's capability to produce hand-written style, a significantly more challenging task due to the considerable differences from typed fonts. 
In Figure~\ref{fig:typefaces} we showcase the generation of characters in \emph{Clerical script}, \emph{Semi-cursive script} and \emph{Cursive script}.
It is revealed that our method can adeptly produce characters in calligraphy form, which are markedly distinct from printed form. This success underscores the model's advanced understanding and flexibility, and makes it a potent tool for generating handwritten character styles for all CJK characters, a task not done by any human calligrapher yet.

\subsection{Zero-shot generation to Chinese Character-inspired Writing Systems}

The cultural and historical influence of Sinosphere has lead to other Chinese Character-inspired writing systems beyond the core CJK (Chinese, Japanese and Korean). 
Notably, one such system is Chu Nom~\cite{omniglot2023chunon}, consisting of complex and newly created characters that adhere to Chinese Character system. Historically, Chu Nom was used along side Chu Han (Chinese Character) to write Vietnamese from the 13th to the 20th century. Thanks to significant efforts towards its revival,~\cite{hannomrcv}, we now have a workable references for Chu Nom in the digital publishing era, including NomNaTong font used in this paper. 
Another example is the Tangut script~\cite{omniglot2023tangut}, created by modelling Chinese Characters loosely and employed to write the now-extinct Tangut language from the 10th to the 15th century.
\begin{figure}[t!]
    \centering
    \includegraphics[width=0.85\columnwidth]{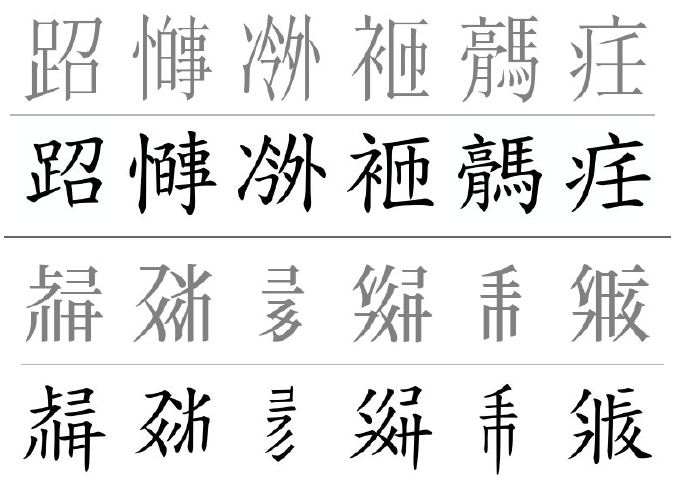}
    \SpaceAfterFigure{}
    \caption{Zero-shot generation of new styles for Chu Nom (upper block) and Tangut scripts (lower block). Gray characters are reference ones and black characters are generated by our method in a new style.}
    \SpaceAfterCaption{}
    \label{fig:chunom-and-tangut}
\end{figure}
A great challenge of such scripts is the extremely limited resource due to lack of active users, as Vietnamese nowadays is written in Latin-based alphabet and Tangut language has no living speakers.
Two fonts in Table~\ref{tbl:fonts} represent almost the entirety of free resources available for these scripts, and there is neither demand nor resources to create fonts in other styles for these writing systems.

To bridge this gap, we apply our proposed method to the setting of zero-shot generation of new styles for these scripts. 
Concretely, our method generate \emph{regular script} based on \emph{Ming/Song} that is trained only with pairs of data in CJK. 
Remarkably, even our proposed method has not seen these out-of-domain characters, it succeeds in producing high-quality results in a new style for both Chu Nom and Tangut scripts.

\subsection{Model Analysis}

\subsection{Comparison with GAN-based Model}

\begin{figure}[h!]
    \centering
    \includegraphics[width=1.0\columnwidth]{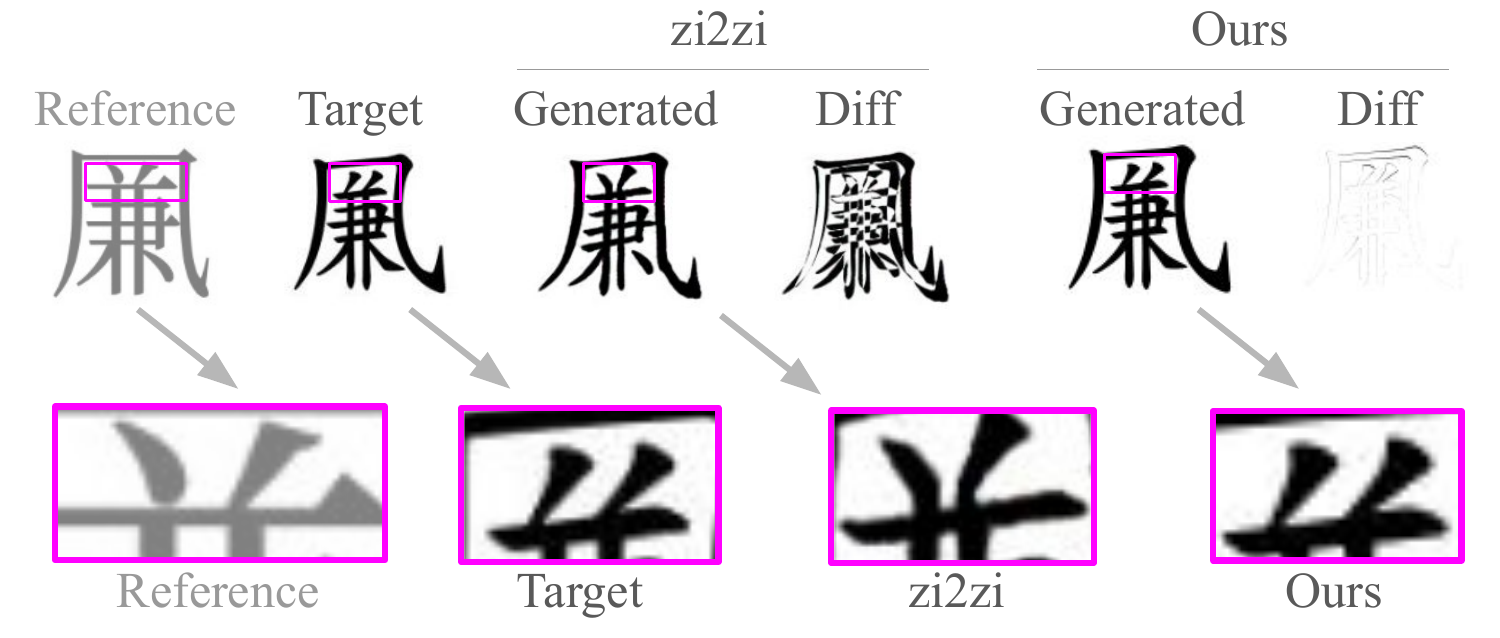}
    \caption{Comparing our method with zi2zi\protect\cite{tian2018zi2zi}. Ours methods generated characters that are much closer to the target (compare two ``Diff'') and visually more smoothing (compare two ``Generated''). Furthermore, in the zoom-ins below show that our method successfully transfers into a more global structure in the target style, while zi2zi fails and instead copies the reference.}
    \SpaceAfterCaption{}
    \label{fig:ours-vs-zi2zi}
\end{figure}

Diffusion models have recently outperformed GAN-based counterparts in many applications, which have long been the predominant approach in generative modeling. 
Our work is a clear example of this paradigm shift.
As shown in Figure~\ref{fig:ours-vs-zi2zi}, our method could generate characters with visually higher quality. 
Crucially, it also succeeds in capturing the more global structure of the desired style, a task for which prior models have failed.

\begin{figure}[h!]
    \centering
    \includegraphics[width=0.85\columnwidth]{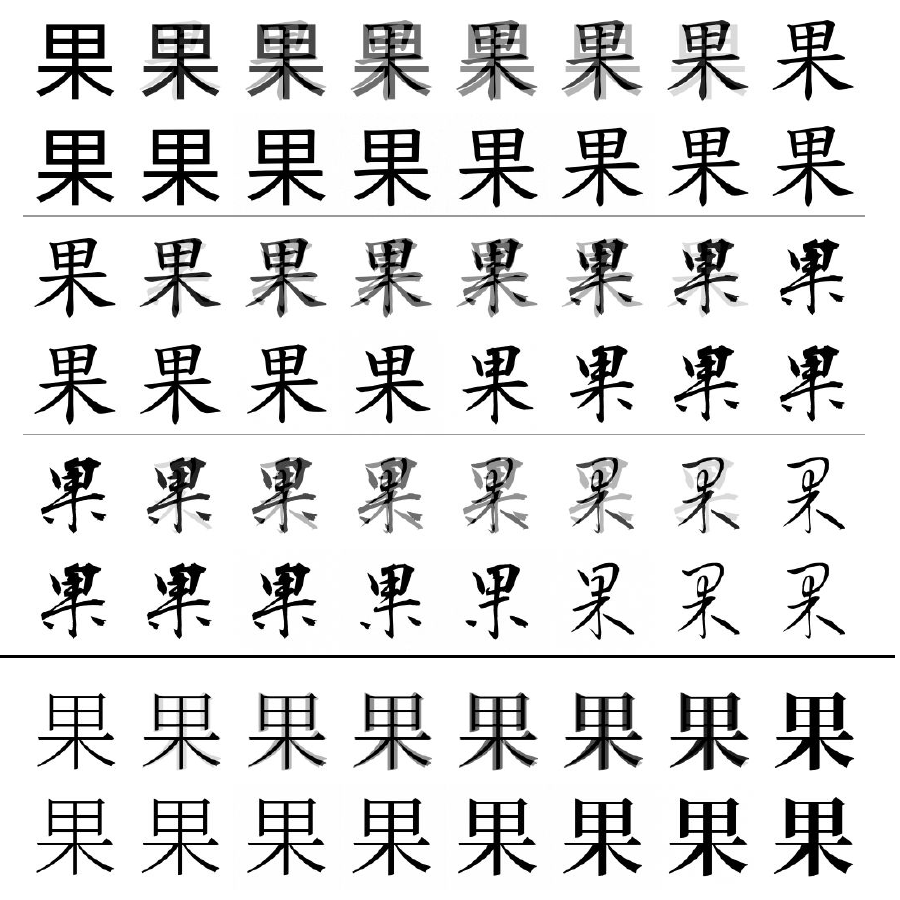}
    \SpaceAfterFigure{}
    \caption{Interpolation of the same character. \textbf{Above}: interpolating from \emph{Gothic typefaces} to \emph{Regular script} (upper block), then to \emph{Semi-cursive script} (middle block), then to \emph{Cursive script} (lower block). \textbf{Lower}: interpolating from \emph{Ultra Thin} to \emph{Bold}. In each block above row shows superficially mixing character at pixel level and the lower row is generated by our method.}
    \SpaceAfterCaption{}
    \label{fig:interpolation}
\end{figure}

\subsubsection{Style Interpolation}
The design of the architecture in our method allows free interpolation between different styles through a weighted mixture of data embeddings.
As we show in Figure~\ref{fig:interpolation}, our method facilitates smooth transitions across different styles (printing or handwriting), and the same for font weight from \emph{Ultra Thin} to \emph{Bold} and the generated intermediate results are coherent and meaningful.
This capability is particularly helpful for creating stylized fonts that blend various styles, offering a painlessly pipeline for all CJK characters.

\begin{figure}[h!]
    \centering
    \includegraphics[width=0.75\columnwidth]{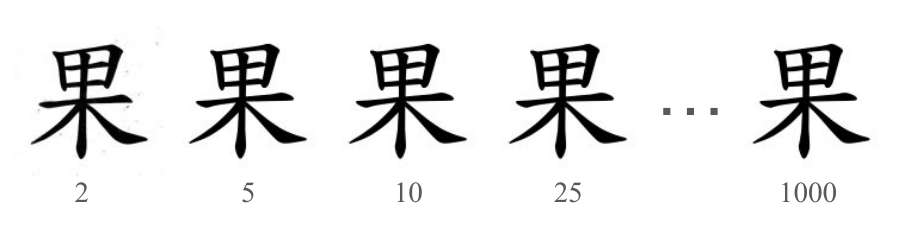}
    \includegraphics[width=0.75\columnwidth]{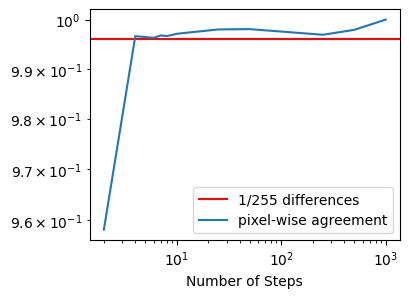}
    \SpaceAfterFigure{}
    \caption{Inference using different number of steps for diffusion. We here show the results with different number of steps. It could be seen that only the most extreme case of two steps we may see a visible artifact. We also compare a wide range of steps by looking at the mean of pixel-wise matching between them and the result from 1000 steps. }
    \SpaceAfterCaption{}
    \label{fig:steps}
\end{figure}

\begin{figure}[h!]
    \centering
    \includegraphics[width=0.8\columnwidth]{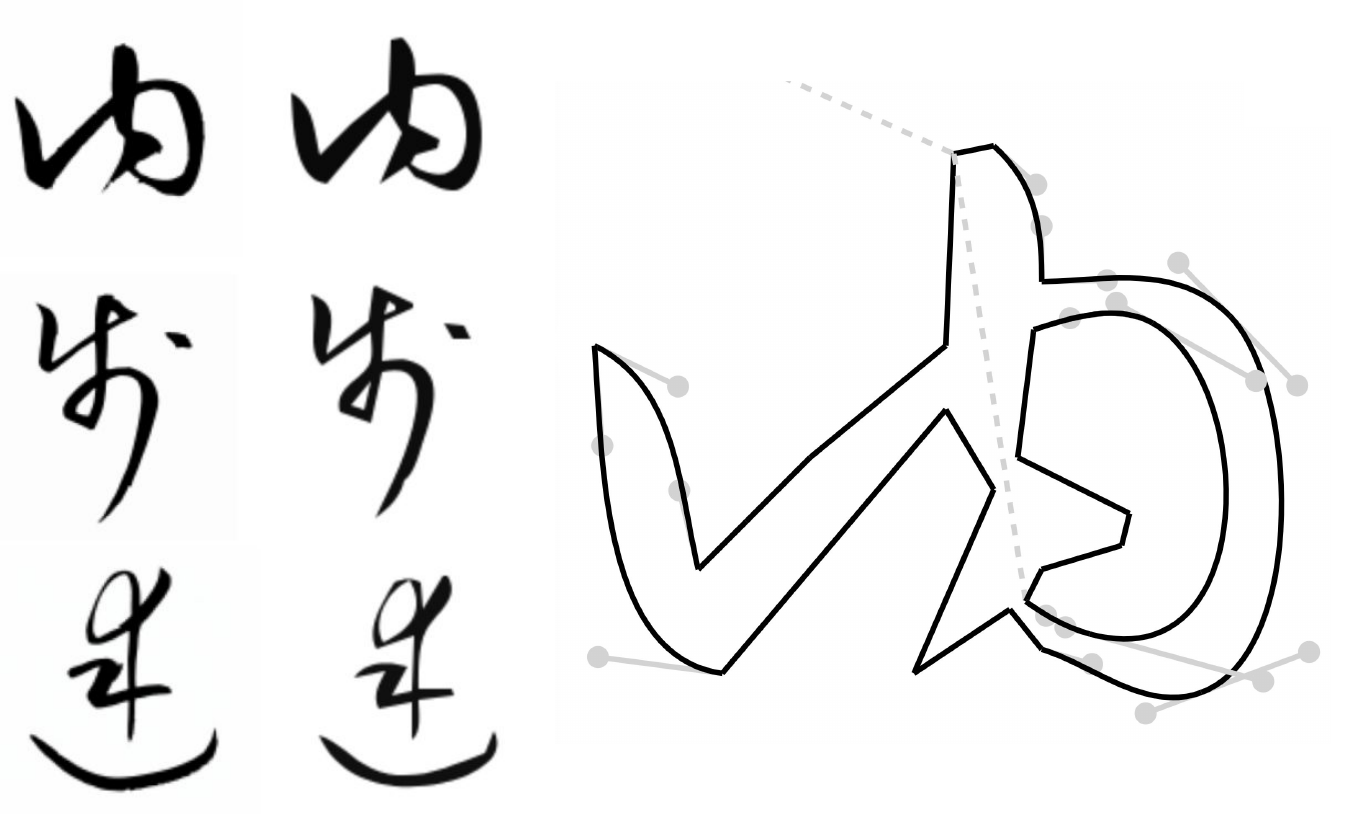}
    \SpaceAfterFigure{}
    \caption{Example of vectorziation. On the left we show two columns: generated images and vectorized SVG files. On the right are the visualized SVG paths that shows control commands, where solid grey lines for Bezier curve and dash grey lines for moving pen. Painted using SVG Path Visualizer~\protect\cite{mathieu2020svg}.}
    \label{fig:example-of-vectorization}
\end{figure}

\subsection{Performance Study: Diffusion Steps v.s. Quality.}

The diffusion model is trained and inferred with discretization of time steps from the continuous timespace $[0,1]$ into $T$ steps. Although we train the model with a discretization of $T=1000$ steps, it is suggested~\cite{lu2022dpm,xue2024sa} that effective inference can be achieved with significantly fewer steps.
While we do not explicitly employ techniques to optimize for few-step sampling, 
exploring the minimum number of steps required for accurate generation remains valuable.
As shown in Figure~\ref{fig:steps},  five steps are sufficient and could be treated as feasible approximation.
This efficiency allows for generating a character in under one second on a T4 GPU, which is considered less powerful and more suited for inference tasks.

\subsection{Vectorization of Generated Characters}

We finally show the vectorization of generated characters in Figure~\ref{fig:example-of-vectorization} using the autotrace tool~\cite{autotrace}.
This process is necessary since our method produces bitmap while a font designer would expect a vectorized glyph. 
The result is satisfactory. 
It's important to note that vectorization parameters are highly sensitive to specific fonts and must be carefully adjusted for each font.

\section{Conclusion}
In this paper, we propose a diffusion based method that generates glyph in a wide range of target style from a single reference glyph.
Our experiments show that our model does well for different styles, making it useful for both font design and artistical purpose.

Future work directions could include (1) deepening the integration into font design pipeline, (2) extending to multiple character typesetting and calligraphy where the interaction \emph{between} the characters could be modeled, and (3) more ambiguously, integrating into powerful text-to-image models to enable accurate label generation.

\section{Acknowledgement}
We thank Marco Raymond Cognetta and Chris Simpkins for their valuable feedback.
\bibliographystyle{iccc}
\bibliography{ref}

\begin{thebibliography}{}

\bibitem[\protect\citeauthoryear{Chiba and others}{2020}]{w3jlreq}
Chiba, H., et~al.
\newblock 2020.
\newblock Requirements for japanese text layout.
\newblock W3c working group note, W3C.
\newblock \url{https://www.w3.org/TR/jlreq/}.

\bibitem[\protect\citeauthoryear{Croitoru \bgroup et al.\egroup }{2023}]{croitoru2023diffusion}
Croitoru, F.-A.; Hondru, V.; Ionescu, R.~T.; and Shah, M.
\newblock 2023.
\newblock Diffusion models in vision: A survey.
\newblock {\em IEEE Transactions on Pattern Analysis and Machine Intelligence}.

\bibitem[\protect\citeauthoryear{Da}{2010}]{jun2010chinese}
Da, J.
\newblock 2010.
\newblock Chinese text computing.
\newblock \url{https://lingua.mtsu.edu/chinese-computing/}.
\newblock Accessed: 2024-01-01.

\bibitem[\protect\citeauthoryear{Dutour}{2020}]{mathieu2020svg}
Dutour, M.
\newblock 2020.
\newblock Svg path visualizer.
\newblock \url{https://svg-path-visualizer.netlify.app/}.
\newblock Accessed: 2024-01-01.

\bibitem[\protect\citeauthoryear{Engel \bgroup et al.\egroup }{2020}]{engel2020ddsp}
Engel, J.; Hantrakul, L.; Gu, C.; and Roberts, A.
\newblock 2020.
\newblock Ddsp: Differentiable digital signal processing.
\newblock {\em arXiv preprint arXiv:2001.04643}.

\bibitem[\protect\citeauthoryear{{Ethnologue Authors}}{2024}]{authors2024ethnologue}
{Ethnologue Authors}.
\newblock 2024.
\newblock Ethnologue: Languages of the world.
\newblock \url{https://www.ethnologue.com/}.
\newblock Accessed: 2024-02-01.

\bibitem[\protect\citeauthoryear{Gao \bgroup et al.\egroup }{2019}]{gao2019artistic}
Gao, Y.; Guo, Y.; Lian, Z.; Tang, Y.; and Xiao, J.
\newblock 2019.
\newblock Artistic glyph image synthesis via one-stage few-shot learning.
\newblock {\em ACM Transactions on Graphics (TOG)} 38(6):1--12.

\bibitem[\protect\citeauthoryear{Goel \bgroup et al.\egroup }{2022}]{goel2022s}
Goel, K.; Gu, A.; Donahue, C.; and R{\'e}, C.
\newblock 2022.
\newblock It’s raw! audio generation with state-space models.
\newblock In {\em International Conference on Machine Learning},  7616--7633.
\newblock PMLR.

\bibitem[\protect\citeauthoryear{Gui \bgroup et al.\egroup }{2023}]{gui2023zero}
Gui, D.; Chen, K.; Ding, H.; and Huo, Q.
\newblock 2023.
\newblock Zero-shot generation of training data with denoising diffusion probabilistic model for handwritten chinese character recognition.
\newblock {\em arXiv preprint arXiv:2305.15660}.

\bibitem[\protect\citeauthoryear{{HNRCV Authors}}{}]{hannomrcv}
{HNRCV Authors}.
\newblock {Han-nom Revival Committe of Vietnam}.
\newblock \url{https://www.hannom-rcv.org/}.

\bibitem[\protect\citeauthoryear{Ho, Jain, and Abbeel}{2020}]{ho2020denoising}
Ho, J.; Jain, A.; and Abbeel, P.
\newblock 2020.
\newblock Denoising diffusion probabilistic models.
\newblock {\em Advances in neural information processing systems} 33:6840--6851.

\bibitem[\protect\citeauthoryear{Huang}{2020}]{huang2020type}
Huang, L.
\newblock 2020.
\newblock Type classification in cjk: Chinese.
\newblock \url{https://fonts.google.com/knowledge/type_in_china_japan_and_korea/type_classification_in_cjk_chinese}.
\newblock Accessed: 2024-01-01.

\bibitem[\protect\citeauthoryear{{IMF}}{2023}]{imf2023world}
{IMF}.
\newblock 2023.
\newblock "world economic outlook database, october 2023".
\newblock \url{https://www.imf.org/en/Publications/WEO/weo-database/2023/October}.
\newblock Accessed: 2024-02-01.

\bibitem[\protect\citeauthoryear{Kamichi}{2023}]{jigmofonts}
Kamichi, K.
\newblock 2023.
\newblock Jigmo（字雲）font.
\newblock \url{https://kamichikoichi.github.io/jigmo/}.
\newblock Accessed: 2024-01-01.

\bibitem[\protect\citeauthoryear{Kim}{2020a}]{kim2020typejp}
Kim, M.-Y.
\newblock 2020a.
\newblock Type classification in cjk: Japanese.
\newblock \url{https://fonts.google.com/knowledge/type_in_china_japan_and_korea/type_classification_in_cjk_japanese}.
\newblock Accessed: 2024-01-01.

\bibitem[\protect\citeauthoryear{Kim}{2020b}]{kim2020typekr}
Kim, M.-Y.
\newblock 2020b.
\newblock Type classification in cjk: Korean.
\newblock \url{https://fonts.google.com/knowledge/type_in_china_japan_and_korea/type_classification_in_cjk_korean}.
\newblock Accessed: 2024-01-01.

\bibitem[\protect\citeauthoryear{Lim and others}{2020}]{w3klreq}
Lim, S.-B., et~al.
\newblock 2020.
\newblock Requirements for hangul text layout and typography.
\newblock W3c working group note, W3C.
\newblock \url{https://www.w3.org/TR/klreq/}.

\bibitem[\protect\citeauthoryear{Lisa~Huang}{2020}]{authors20xxtype}
Lisa~Huang, M.-Y.~K.
\newblock 2020.
\newblock Type in china, japan, and korea.
\newblock \url{https://fonts.google.com/knowledge/type_in_china_japan_and_korea}.
\newblock Accessed: 2024-01-01.

\bibitem[\protect\citeauthoryear{Liu and Lian}{2023}]{liu2023fonttransformer}
Liu, Y., and Lian, Z.
\newblock 2023.
\newblock Fonttransformer: Few-shot high-resolution chinese glyph image synthesis via stacked transformers.
\newblock {\em Pattern Recognition} 141:109593.

\bibitem[\protect\citeauthoryear{Liu}{2022}]{liu2022deciphering}
Liu, Y.
\newblock 2022.
\newblock Deciphering the hanzisphere.
\newblock \url{https://www.sixthtone.com/news/1011502}.
\newblock Accessed: 2024-01-01.

\bibitem[\protect\citeauthoryear{Lovelace \bgroup et al.\egroup }{2024}]{lovelace2024latent}
Lovelace, J.; Kishore, V.; Wan, C.; Shekhtman, E.; and Weinberger, K.~Q.
\newblock 2024.
\newblock Latent diffusion for language generation.
\newblock {\em Advances in Neural Information Processing Systems} 36.

\bibitem[\protect\citeauthoryear{Lu \bgroup et al.\egroup }{2022}]{lu2022dpm}
Lu, C.; Zhou, Y.; Bao, F.; Chen, J.; Li, C.; and Zhu, J.
\newblock 2022.
\newblock Dpm-solver: A fast ode solver for diffusion probabilistic model sampling in around 10 steps.
\newblock {\em Advances in Neural Information Processing Systems} 35:5775--5787.

\bibitem[\protect\citeauthoryear{Marginson}{2011}]{marginson2011higher}
Marginson, S.
\newblock 2011.
\newblock Higher education in east asia and singapore: Rise of the confucian model.
\newblock {\em Higher education} 61:587--611.

\bibitem[\protect\citeauthoryear{{MidJourney Authors}}{2024}]{authors2024midjourney}
{MidJourney Authors}.
\newblock 2024.
\newblock Midjourney.
\newblock \url{https://www.midjourney.com/home}.
\newblock Accessed: 2024-02-09.

\bibitem[\protect\citeauthoryear{Needham}{1974}]{needham1974science}
Needham, J.
\newblock 1974.
\newblock {\em Science and civilisation in China}, volume~5.
\newblock Cambridge University Press.

\bibitem[\protect\citeauthoryear{Nichol and Dhariwal}{2021}]{nichol2021improved}
Nichol, A.~Q., and Dhariwal, P.
\newblock 2021.
\newblock Improved denoising diffusion probabilistic models.
\newblock In {\em International Conference on Machine Learning},  8162--8171.
\newblock PMLR.

\bibitem[\protect\citeauthoryear{NINJAL}{2015}]{ninjal2015}
NINJAL.
\newblock 2015.
\newblock 『現代日本語書き言葉均衡コーパス』語彙表 (vocabulary list for ``the balanced corpus of contemporary written japanese'').
\newblock \url{https://clrd.ninjal.ac.jp/bccwj/freq-list.html}.
\newblock Accessed: 2024-01-01.

\bibitem[\protect\citeauthoryear{{Noto Authors}}{2020}]{notofonts}
{Noto Authors}.
\newblock 2020.
\newblock Noto fonts.
\newblock \url{https://fonts.google.com/noto/use}.
\newblock Accessed: 2024-01-01.

\bibitem[\protect\citeauthoryear{omniglot}{2023a}]{omniglot2023chunon}
omniglot.
\newblock 2023a.
\newblock Chu-non script.
\newblock \url{https://www.omniglot.com/writing/chunom.htm}.
\newblock Accessed: 2024-01-01.

\bibitem[\protect\citeauthoryear{omniglot}{2023b}]{omniglot2023tangut}
omniglot.
\newblock 2023b.
\newblock Tangut.
\newblock \url{https://www.omniglot.com/writing/tangut.htm}.
\newblock Accessed: 2024-01-01.

\bibitem[\protect\citeauthoryear{Park \bgroup et al.\egroup }{2021}]{park2021few}
Park, S.; Chun, S.; Cha, J.; Lee, B.; and Shim, H.
\newblock 2021.
\newblock Few-shot font generation with localized style representations and factorization.
\newblock In {\em Proceedings of the AAAI conference on artificial intelligence}, volume~35,  2393--2402.

\bibitem[\protect\citeauthoryear{Podell \bgroup et al.\egroup }{2023}]{podell2023sdxl}
Podell, D.; English, Z.; Lacey, K.; Blattmann, A.; Dockhorn, T.; M{\"u}ller, J.; Penna, J.; and Rombach, R.
\newblock 2023.
\newblock Sdxl: Improving latent diffusion models for high-resolution image synthesis.
\newblock {\em arXiv preprint arXiv:2307.01952}.

\bibitem[\protect\citeauthoryear{Ronneberger, Fischer, and Brox}{2015}]{ronneberger2015u}
Ronneberger, O.; Fischer, P.; and Brox, T.
\newblock 2015.
\newblock U-net: Convolutional networks for biomedical image segmentation.
\newblock In {\em Medical Image Computing and Computer-Assisted Intervention--MICCAI 2015: 18th International Conference, Munich, Germany, October 5-9, 2015, Proceedings, Part III 18},  234--241.
\newblock Springer.

\bibitem[\protect\citeauthoryear{Ruiz \bgroup et al.\egroup }{2023}]{ruiz2023dreambooth}
Ruiz, N.; Li, Y.; Jampani, V.; Pritch, Y.; Rubinstein, M.; and Aberman, K.
\newblock 2023.
\newblock Dreambooth: Fine tuning text-to-image diffusion models for subject-driven generation.
\newblock In {\em Proceedings of the IEEE/CVF Conference on Computer Vision and Pattern Recognition},  22500--22510.

\bibitem[\protect\citeauthoryear{Sohl-Dickstein \bgroup et al.\egroup }{2015}]{sohl2015deep}
Sohl-Dickstein, J.; Weiss, E.; Maheswaranathan, N.; and Ganguli, S.
\newblock 2015.
\newblock Deep unsupervised learning using nonequilibrium thermodynamics.
\newblock In {\em International conference on machine learning},  2256--2265.
\newblock PMLR.

\bibitem[\protect\citeauthoryear{Song \bgroup et al.\egroup }{2020}]{song2020score}
Song, Y.; Sohl-Dickstein, J.; Kingma, D.~P.; Kumar, A.; Ermon, S.; and Poole, B.
\newblock 2020.
\newblock Score-based generative modeling through stochastic differential equations.
\newblock {\em arXiv preprint arXiv:2011.13456}.

\bibitem[\protect\citeauthoryear{Stocks}{2020a}]{elliot2020introducing}
Stocks, E.~J.
\newblock 2020a.
\newblock Introducing weights \& styles.
\newblock \url{https://fonts.google.com/knowledge/introducing_type/introducing_weights_styles}.
\newblock Accessed: 2024-01-01.

\bibitem[\protect\citeauthoryear{Stocks}{2020b}]{elliot2020making}
Stocks, E.~J.
\newblock 2020b.
\newblock Making sense of typographic classifications.
\newblock \url{https://fonts.google.com/knowledge/introducing_type/making_sense_of_typographic_classifications}.
\newblock Accessed: 2024-01-01.

\bibitem[\protect\citeauthoryear{Tian}{2018}]{tian2018zi2zi}
Tian, Y.
\newblock 2018.
\newblock zi2zi: Master chinese calligraphy with conditional adversarial networks.
\newblock \url{https://github.com/kaonashi-tyc/zi2zi}.
\newblock Accessed: 2024-01-01.

\bibitem[\protect\citeauthoryear{Tung and others}{2023}]{w3clreq}
Tung, B., et~al.
\newblock 2023.
\newblock Requirements for chinese text layout.
\newblock W3c group draft note, W3C.
\newblock \url{https://www.w3.org/TR/clreq/}.

\bibitem[\protect\citeauthoryear{Unicode}{2023}]{unicode2023v15dot1}
Unicode.
\newblock 2023.
\newblock Unicode 15.1 character code charts.
\newblock \url{https://unicode.org/charts/}.
\newblock Accessed: 2024-01-01.

\bibitem[\protect\citeauthoryear{Whistler}{2023}]{unicodetechnicalnote26}
Whistler, K.
\newblock 2023.
\newblock On the encoding of latin, greek, cyrillic and han.
\newblock Unicode technical note, Unicode Consortium.
\newblock \url{https://www.unicode.org/notes/tn26/}.

\bibitem[\protect\citeauthoryear{{Wikipedia contributors}}{2024}]{wiki:commonsancient}
{Wikipedia contributors}.
\newblock 2024.
\newblock Wikimedia commons:ancient chinese characters project.
\newblock Accessed: 2024-01-01.

\bibitem[\protect\citeauthoryear{Wu \bgroup et al.\egroup }{2024}]{wu2024ar}
Wu, T.; Fan, Z.; Liu, X.; Zheng, H.-T.; Gong, Y.; Jiao, J.; Li, J.; Guo, J.; Duan, N.; Chen, W.; et~al.
\newblock 2024.
\newblock Ar-diffusion: Auto-regressive diffusion model for text generation.
\newblock {\em Advances in Neural Information Processing Systems} 36.

\bibitem[\protect\citeauthoryear{Xue \bgroup et al.\egroup }{2024}]{xue2024sa}
Xue, S.; Yi, M.; Luo, W.; Zhang, S.; Sun, J.; Li, Z.; and Ma, Z.-M.
\newblock 2024.
\newblock Sa-solver: Stochastic adams solver for fast sampling of diffusion models.
\newblock {\em Advances in Neural Information Processing Systems} 36.

\bibitem[\protect\citeauthoryear{Yamato \bgroup et al.\egroup }{2020}]{autotrace}
Yamato, M.; Lemenkov, P.; Weber, M.; et~al.
\newblock 2020.
\newblock autotrace.
\newblock \url{https://github.com/autotrace/autotrace}.
\newblock Accessed: 2024-01-01.

\bibitem[\protect\citeauthoryear{Yang \bgroup et al.\egroup }{2023}]{yang2023diffusion}
Yang, L.; Zhang, Z.; Song, Y.; Hong, S.; Xu, R.; Zhao, Y.; Zhang, W.; Cui, B.; and Yang, M.-H.
\newblock 2023.
\newblock Diffusion models: A comprehensive survey of methods and applications.
\newblock {\em ACM Computing Surveys} 56(4):1--39.

\bibitem[\protect\citeauthoryear{Yun-Chen~Lo}{2019}]{lo2019font2font}
Yun-Chen~Lo, Yi-Ren~Chen, J. Z.~J.
\newblock 2019.
\newblock Font2font.
\newblock \url{https://github.com/yunchenlo/Font2Font}.
\newblock Accessed: 2024-01-01.

\bibitem[\protect\citeauthoryear{Zhang, Zhang, and Cai}{2018}]{zhang2018separating}
Zhang, Y.; Zhang, Y.; and Cai, W.
\newblock 2018.
\newblock Separating style and content for generalized style transfer.
\newblock In {\em Proceedings of the IEEE conference on computer vision and pattern recognition},  8447--8455.

\bibitem[\protect\citeauthoryear{Zhou \bgroup et al.\egroup }{2023}]{zhou2023shifted}
Zhou, Y.; Liu, B.; Zhu, Y.; Yang, X.; Chen, C.; and Xu, J.
\newblock 2023.
\newblock Shifted diffusion for text-to-image generation.
\newblock In {\em Proceedings of the IEEE/CVF Conference on Computer Vision and Pattern Recognition},  10157--10166.

\bibitem[\protect\citeauthoryear{Zhu \bgroup et al.\egroup }{2020}]{zhu2020few}
Zhu, A.; Lu, X.; Bai, X.; Uchida, S.; Iwana, B.~K.; and Xiong, S.
\newblock 2020.
\newblock Few-shot text style transfer via deep feature similarity.
\newblock {\em IEEE Transactions on Image Processing} 29:6932--6946.

\end{thebibliography}
\end{CJK}
\end{document}